%% file: Su_2020_trb.tex
\documentclass[numbered]{trbunofficial}
\usepackage{graphicx}
\usepackage{algorithm}
\usepackage{algorithmic}
\usepackage{subcaption}
\usepackage{mwe}
\usepackage{tikz}
\usepackage{pgfplots}
\usepackage{placeins}
\usepackage{mfirstuc}
\pgfplotsset{compat=1.14}

\usepackage[hidelinks]{hyperref}
\DeclareMathOperator*{\argmax}{arg\,max}
\AuthorHeaders{Su, Shi, Jin, Chow}
\title{V2I Connectivity-Based Dynamic Queue-Jump Lane for Emergency Vehicles: A Deep Reinforcement Learning Approach}

\author{%
  \textbf{Haoran Su}\\
  Department of Civil and Urban Engineering\\
  C2SMART University Transportation Center\\
  New York University Tandon School of Engineering\\
  15 MetroTech Center, Brooklyn, NY 11201, USA\\
  haoran.su@nyu.edu\\
  \hfill\break
  \textbf{Kejian Shi}\\
  Department of Computer Science and Engineering\\
  New York University Tandon School of Engineering\\
  370 Jay Street, Brooklyn, NY 11201, USA\\
  ks4765@nyu.edu\\
  \hfill\break%
  \textbf{Li Jin, Ph.D.}\\
  Department of Civil and Urban Engineering\\
  C2SMART University Transportation Center\\
  New York University Tandon School of Engineering\\
  15 MetroTech Center, Brooklyn, NY 11201, USA\\
  lijin@nyu.edu\\
  \hfill\break%
  \textbf{Joseph Y.J. Chow, Ph.D.}\\
  Department of Civil and Urban Engineering\\
  C2SMART University Transportation Center\\
  New York University Tandon School of Engineering\\
  15 MetroTech Center, Brooklyn, NY 11201, USA\\
  joseph.chow@nyu.edu\\
  \hfill\break%
}
\TotalWords{5299}

\begin{document}

\maketitle

\section*{Abstract}
Emergency vehicle (EMV) service is a key function of cities and is exceedingly challenging due to urban traffic congestion. A main reason behind EMV service delay is the lack of communication and cooperation between vehicles blocking EMVs. In this paper, we study the improvement of EMV service under V2I connectivity.
We consider the establishment of dynamic queue jump lanes (DQJLs) based on real-time coordination of connected vehicles. We develop a novel Markov decision process formulation for the DQJL problem, which explicitly accounts for the uncertainty of drivers' reaction to approaching EMVs. We propose a deep neural network-based reinforcement learning algorithm that efficiently computes the optimal coordination instructions. We also validate our approach on a micro-simulation testbed using Simulation of Urban Mobility (SUMO). Validation results show that with our proposed methodology, the centralized control system saves approximately 15\% EMV passing time than the benchmark system.

\hfill\break%
\noindent\textit{Keywords}: Deep Reinforcement Learning, Connected Vehicles, Emergency Vehicles
\newpage
\input{./text/10_Intro}
\input{./text/20_Modeling}
\input{./text/30_Methodology}

\input{./text/40_Simulation}
\input{./text/50_Conclusion}

\newpage
\bibliographystyle{trb}
\bibliography{Su_2020_trb}
\end{document}

%% file: text/10_Intro.tex
\section{Introduction}

Increasing population and urbanization have made it exceedingly challenging to operate urban emergency services efficiently. For example, historical data from New York City, USA \cite{NY2019} shows that the number of annual emergency vehicle (EMV) incidents has grown from 1,114,693 in 2004 to 1,352,766 in 2014, with corresponding average response times of 7:53 min and 9:23 min, respectively \cite{Emergency2014}. This means an approximately 20\% increase in response times in ten years. In the case of cardiac arrest, every minute until defibrillation reduces survival chances by 7\% to 10\%, and after 8 minutes there is little chance for survival \cite{Heart2013}. Cities are less resilient with worsening response times from EMVs (ambulances, fire trucks, police cars), mainly due to traffic congestion.

In recent years, machine learning based models have achieved remarkable success in a wide range of application areas \cite{you2020unsupervised,you2020data,you2021knowledge,chen2021adaptive,you2020contextualized,you2018structurally,you2019low,you2019ct,you2021momentum}. The performance of these EMV service systems in congested traffic can be improved with technology. 
As a core of modern ITSs, wireless vehicle-to-infrastructure (V2I) connectivity, such as Cellular Networks, provide significant opportunities for improving urban emergency response. On the one hand, wireless connectivity provides EMVs the traffic conditions on possible routes between the station (hospital, fire station, police station, etc.) and the call, which enables more efficient dispatch and routing. On the other hand, through V2I communications, traffic managers can broadcast the planned route of EMVs to non-EMVs that may be affected, and non-EMVs can cooperate to form dynamic queue-jump lanes (QJLs) for approaching EMVs.

Queue jump lane (DQJ) introduced by \cite{Zhou2005Performance, Cesme2015Queue, Farid2015Dedicated}, utilizes an additional static lane for bus operation so that buses can bypass long queues before intersections. It has never been used for EMV deployment and is considered a novel operation strategy to apply this technology for EMV deployment with the aid of connected vehicle technologies.
Although QJLs are a relatively new technology, literature is available documenting the positive effects they have in reducing travel time variability, especially when used in conjunction with the transit signal priority (TSP). However, they are all based on moving-bottleneck models for buses \cite{Zhou2005Performance,Cesme2015Queue,Farid2015Dedicated}; we are borrowing this bus operation strategy for EMV deployment in our setting, since EMVs typically move faster than non-EMVs and since EMVs can “preempt” non-EMV traffic because of their priority. 
On the other hand, with siren technologies, other vehicles often do not get enough warning time from the EMVs. Even then, there is a lack of clarity in the direction of the route to avoid, not to mention the added noise pollution. The confusion, particularly under highly congested scenarios, leads to increased delays as mentioned above, and also to 4 to 17 times higher accident rates \cite{Buchenscheit2009AVE} and increased severity of collisions \cite{Yasmin2012Effects}, which further lead to increased response times. A study by Savolainen et al. \cite{Savolainen2010Effects} using data from Detroit, Michigan, verified this sensitivity of driver behavior to different ITS communication methods for EMV route information. 

\begin{figure}[ht]
    \centering
    \begin{subfigure}{\textwidth}  
        \centering 
        \includegraphics[width=0.85\textwidth]{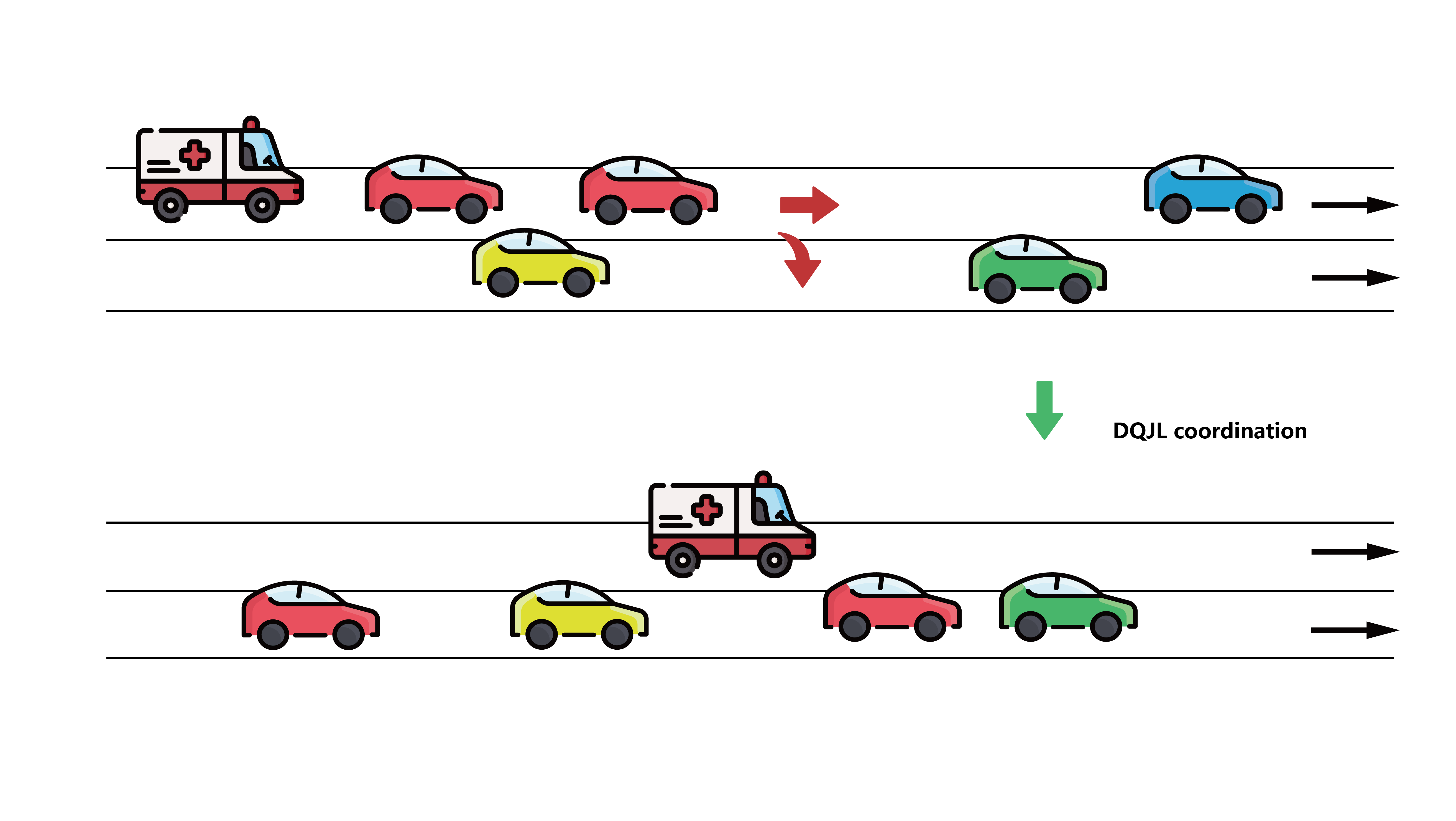}
        \caption{Establishing a DQJL for Emergency Vehicle.}
        \label{fig:dqjl_demo}
    \end{subfigure}
    \hfill
    \begin{subfigure}{\textwidth}
        \centering
        \includegraphics[width=\textwidth]{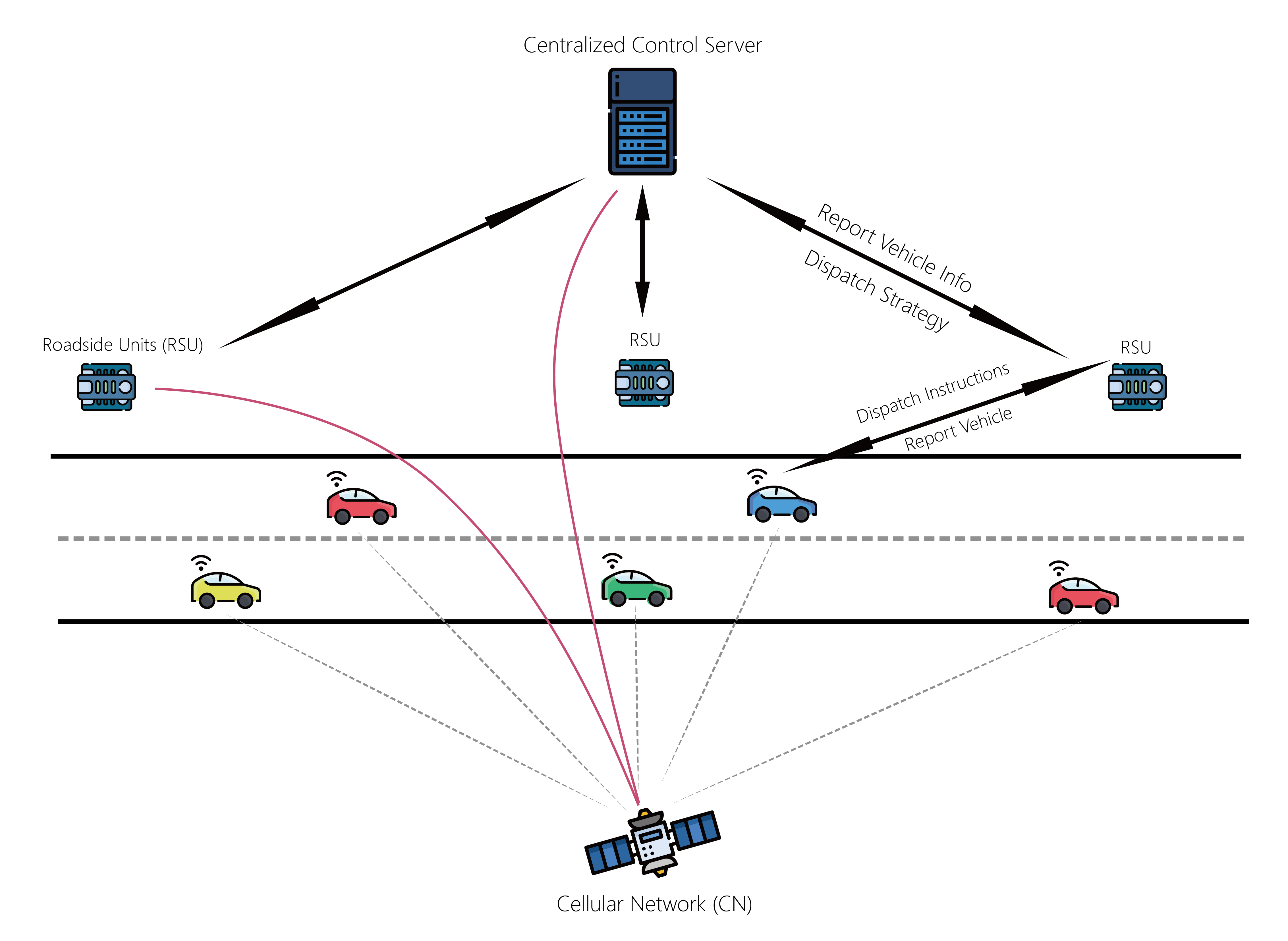}
        \caption{The dynamic queue-jump lane (DQJL) coordination framework via cellular network (CN) and connected vehicles technologies.}    
        \label{fig:CN_framework}
    \end{subfigure}
    \caption{V2I DQJL coordination framework.}
    \label{fig:DQJL_framework}
\end{figure}

In addition, QJLs have not been studied as a dynamic control strategy and there's an urgent need for capturing uncertainties in realistic traffic conditions, especially in events under non-deterministic setting such as yielding for an approaching EMV. To apply dynamic control strategy for vehicle's motion planning for QJLs establishment, we introduce the concept of dynamic queue jump lane (DQJL), see Figure 
\ref{fig:dqjl_demo}. During the process of clearing an QJL for the passing EMV, non-EMVs are constantly monitored and instructed for actions so that DQJL can be established quickly with safety.
\FloatBarrier
The establishment of DQJLs involves real-time motion planning for vehicles, which has been a focus of robotics both in deterministic and in stochastic settings \cite{Buchenscheit2009AVE}. However, although robotic motion planning algorithms provide useful insights for EMVs, they do not directly apply to EMVs, since human drivers are not able to follow high frequency instructions and react instantaneously as robots do. Furthermore, coordination algorithms for multiple robots are hardly applicable to traffic management due to high randomness in drivers’ reaction to coordination instructions. Instead, human drivers need driving strategies that are easy to interpret and implement and preferably only dependent of movement of neighboring vehicles, see \cite{Krajzewicz2002b}. A study by Zuo et al. \cite{Fan2019Microscopic} illustrates to use dynamic programming to prevent vehicle-passenger collision and \cite{Xiong2016CombiningDR} shows how to employ reinforcement learning methods to ensure vehicle safety. Mixed integer programming has been utilized in routing problems for multiple vehicles in different tasks like in \cite{Schouwenaars2001Mixed}. In particular, \cite{Hannoun2019Facilitating} considered an integer linear program formulation for non-EMVs arrangement of a discrete road segment with approaching EMV in the centralized and deterministic setting, which provides a baseline but does not account for the randomness of driver behavior as well as unique attributes road environment such as vehicle length and deceleration profiles. More importantly, it doesn't consider the real-time characteristic and loses practicality in the DQJL application.

In response to these challenges, this paper develops a methodology for utilizing V2I connectivity to improve EMV services. We design link-level coordination strategies for non-EMVs to fast establish dynamic queue jump lanes (DQJLs) for EMVs while maintaining safety. 
We model the DQJL problem into a Markov decision process to cope with the uncertainty in drivers' behavior and introduce a deep reinforcement learning (RL) model to address the randomness in traffic conditions. Our approach delivers real-time easy-to-follow instructions for human drivers during the DQJL establishment process and achieves application purpose within the shortest amount of time with safety. 
We validate our methodology based on traffic simulation software against benchmark system. We also perform sensitivity analysis to study, under what circumstances, the proposed methodology would prevail.


The rest of this paper is organized as follows. In Section \ref{sec:modeling}, we model the establishment of DQJL under the V2I framework. In Section \ref{sec:methodology} we propose the modified DQN algorithm to solve DQJL problem. In Section \ref{sec:simulation}, the results are validated on validation-based simulation in comparison with benchmark system and sensitivity analysis is conducted to inspect the impacts by different factors. Acknowledgements and author contribution are presented in the Section \ref{sec:acknowledgements} and \ref{sec:author} respectively.

%% file: text/20_Modeling.tex
\section{Model and Problem Statement}\label{sec:modeling}
In this section, we elaborate how we formulate the dynamic queue-jump lane (DQJL) problem and model it in a Markov decision process (MDP) framework.

In order to model the establishment of DQJL for an emergency vehicle (EMV), we can take a look at a typical urban road segment. The urban road segment consists of two lanes facing the same direction. When an EMV on duty is approaching this road segment, the centralized control system based on V2I technologies will send out real-time instructions to all non-EMVs on this road segment. Specifically, one or more roadside units (RSU) will collect real-time information from all vehicles. With the cellular networks (CN) prevailing in V2I communication technologies \cite{SANTA20082850, MILANES201085, MUHAMMAD201850} nowadays, vehicles' information including positions, velocity, acceleration/deceleration and vehicle unique attributes such as vehicle length and most comfortable deceleration will be collected by the RSUs. With CN technologies equipped by all of the connected vehicles, the ad-hoc RSUs are viable and effective in such settings. The RSUs then transmit all data via CNs to the server, i.e. the centralized control system. The centralized control system will process the data and gives optimal coordination strategies instantly, and send back to RSUs. Then, the RSUs broadcast each instruction to the corresponding vehicle and, after a short time period, recollects the data. The cycle repeats until a DQJL is established in this segment. The data transmitting latency is negligible compared with the time horizon for establishing a DQJL, and the server also yields strategy in real-time, see Figure \ref{fig:CN_framework}.

With CNs technologies in the V2I setting, human drivers are practically receiving no more than one instruction from the centralized control system. The instruction is simply yielding or continuing the driver's original trip, so it's easy for all drivers to follow guidance. Rather than using traditional sirens on EMVs, DQJL is applicable to settings where siren cannot; they can go further downstream, go beyond sight distance such as turning in routes, and direct the message to vehicles much more clearly because sirens are ambiguous as they aim at all nearby vehicles rather than specific drivers.

Vehicles who have received the yielding instruction will start to yield, after some perception reaction time, and keep yielding until they park. Vehicles, which are advised to continue, do not receive any instructions from RSUs and they continue their trip to leave this segment. Under such setting, even though the centralized control system is collecting, processing and broadcasting data at a very high frequency, usually 2 to 5 Hz, human drivers will only receive one or no instruction in the coordination process.

\begin{figure}[!ht]
  \centering
  \includegraphics[width=\textwidth]{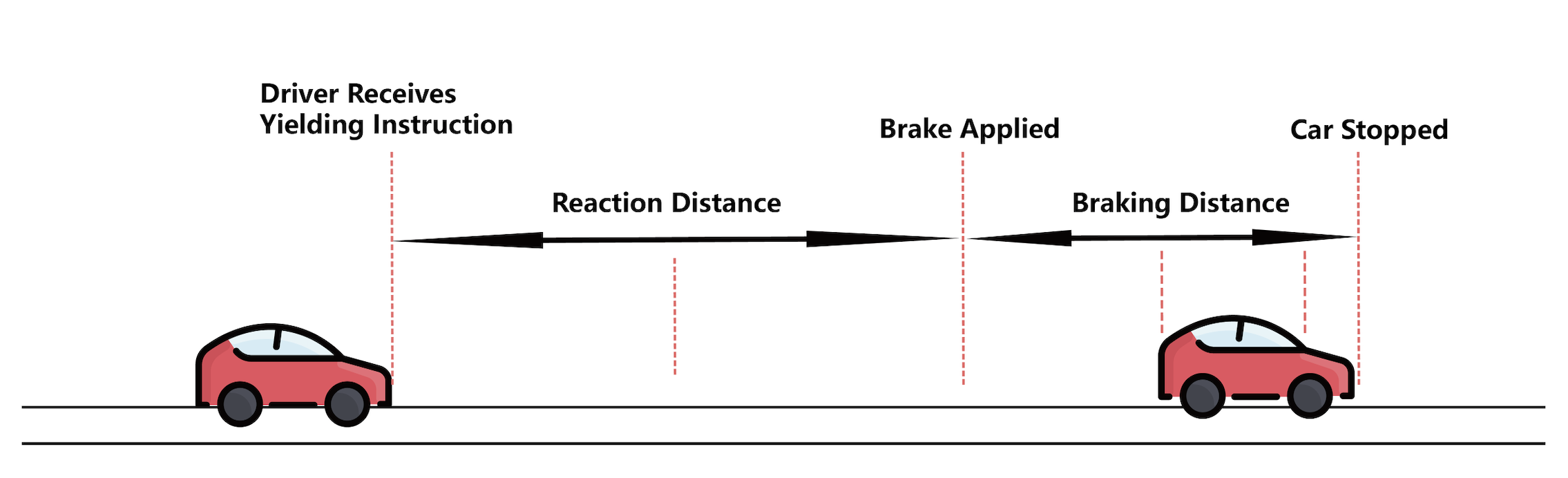}
  \caption{Braking process of a non-EMV after receiving a yielding instruction.}
  \label{fig:braking}
\end{figure}

However, there is uncertainty associated with human drivers' driving behavior, especially when vehicles are yielding. In this study, the uncertainty has two main categories. The first type of uncertainty involves a human driver's perception reaction time. McGehee et al. \cite{McGehee2000} have suggested the human driver's average reaction perception time is 2.3 seconds, but the reaction perception time varies from person to person. The second type of uncertainty is the deceleration adopted by human driver when yielding. Although human drivers will attempt to slow down at the vehicle's most comfortable deceleration, the actual deceleration reflected by vehicle's trajectory varies along time. These two uncertainties will result in variance of vehicle's stopping distance, see Figure \ref{fig:braking}. The variance is generally larger in the pulling-over procedure which includes slowing down and changing lane. These stochastic driving behaviors should be taken into account by the centralized control system when planning the optimal coordination strategy for establishing a DQJL.

\FloatBarrier

\subsection{Problem Statement}
Given a 2-lane road segment facing same direction with unknown number of vehicles, how should the centralized control system formulate a policy to instruct whether and when non-EMVs to yield in real-time, taking into account of human drivers' driving uncertainties, so that a DQJL can be established within the shortest amount of time.

\subsection{Model Assumptions}
There are a few assumptions regarding this modeling: 1. we are only studying a finite urban roadway segment with two lanes; 2. when yielding for an EMV, all non-EMVs' speeds cannot exceed their original speed for safety concerns. They either continue their trips with their original speeds, or be instructed to yield for the EMV and they cannot pull back in the process; 3. when vehicles leave the segment of study, they are considered no longer interacting with other vehicles; 4. during the DQJL process, no new vehicles enter into the segment of study; 5. all vehicles are equipped with CN devices and the CN network communicates data with negligible latency and there is zero packet loss in data transmitting.

\subsection{Markov Decision Process (MDP) Framework}
Due to the uncertainty mentioned above, it is beneficial to picture the controlled system in a Markov Decision Process (MDP) framework so that these randomness can be addressed properly. Notations for variables describing the environment are listed in Table \ref{tab:MDP_notations}.
\begin{table}[!ht]
	\caption{Roadway Environment Notations}\label{tab:MDP_notations}
	\begin{center}
		\begin{tabular}{l l l l}
			Notation & Meaning \\\hline
			$x$                              & front position of the vehicle  \\
			$v$                              & velocity of the vehicle\\
			$l_i$                              & length of the vehicle\\
			$b^{*}$                         & most comfortable acceleration/deceleration of the vehicle\\
			$y$                           & lane position of the vehicle \\
			$L$     & Length of the road segment\\
			$d$     & Minimum safety gap between two vehicles\\
			\hline
		\end{tabular}
	\end{center}
\end{table}

\subsection{State}
The centralized control system considers all non-EMVs as a collection. The positions of all vehicles' fronts, their lane positions, and current velocities at time point/step $t$ are reflected in the state vector $S(t)$. An additional status indicator $z_{i}$ is used to represent if the vehicle is yielding. Namely, the state can be expressed as
\begin{equation*}
S(t) = 
\begin{bmatrix}
x_0(t) & y_0(t) & v_{0}(t) & z_{0}(t)\\
x_1(t) & y_1(t) & v_{1}(t) & z_{1}(t)\\
& \cdots \\
x_{n-1}(t) & y_{n-1}(t) & v_{n-1}(t) & z_{n-1}(t)\\
\end{bmatrix}
.
\end{equation*}

\subsection{Action}
The action for the collection of non-EMVs at step $t$ contains actions instructed by the centralized control system to each vehicle at current time point, which can be described as
\begin{equation*}
    A(t) = [A_{0}(t), A_{1}(t), A_{2}(t), \dots, A_{n-1}(t)],
\end{equation*}
where $A_{i}(t)$ is a binary variable standing for the instruction for the $i$th vehicle at step $t$ is to move forward (0), or yield (1). Non-EMVs on the upper lane receiving a yield instruction will perform a pull-over, whereas those on the lower lane will brake until stop.

\subsection{State Transition}
Once a non-EMV is instructed to yield, the instruction for this vehicle remains to be yielding until the process completes. The status indicator will be consistent with the action value at the current step. To be specific, $z_{i}(t) = A_{i}(t)$ and if $z_{i}(t) = 1, z_{i}(t + m) = 1$, where $m$ is a positive integer. We have
\begin{align*}
    z_{i}(t) = g(z_{i}(t), A_{i}(t)) \begin{cases}
    1, & \text{if } A_{i}(t) = 1 \text{ or } z_{i}(t) = 1,\\
    0, & \text{otherwise}.\\
    \end{cases}
\end{align*}

Combining the driving behavior uncertainty of the perception reaction and the actual deceleration together, we can picture the deceleration as a piece-wise function with respect to step number as
\begin{align*}
    b_{i}(t) = \begin{cases}
    0, & \text{if } A_{i}(t) = 0,\\
    0, & \text{if } A_{i}(t) = 1 \text{ and } t \in [T_{i}, T_{i} + t_{r}),\\
    \hat{b}_{i} + \epsilon_{j}, & \text{ otherwise}.\\
\end{cases}
\end{align*}

In this piece-wise function, $T_{i}$ represents the step in which the vehicle receives a yielding instruction, $t_{r}$ represents the reaction perception time for this driver, $\hat{b}_{i}$ stands for the most comfortable deceleration of the vehicle, $\epsilon_{j}$ is a white noise aiming to capture the deceleration uncertainty. $\epsilon_{j}$ is also a function of $y_{i}(t)$, saying that the deceleration in performing a pull-over and braking until stop are following normal distributions with different standard deviation.

Furthermore, $t_{r}$ can be pictured by a geometric distribution. If the average reaction perception time is 2.3 seconds, it is easy to derive that the probability of success in any step $p = \frac{1}{2.3s/\Delta t} \Rightarrow p = \frac{\Delta t}{2.3s}$, where $\Delta t$ represents the temporal step length.

We also need to determine when a vehicle has successfully pulled over at the end of the step and this can be achieved by examining the $z_{i}(t)$ and $v_{i}(t)$ for non-EMVs with $y_i(t) = 0$:
\begin{align*}
    y_{i}(t+1) = f(y_{i}(t)) = \begin{cases}
        1, &\text{if } z_{i}(t) = 1 \text{ and } v_{i}(t) = 0,\\
        0, &\text{otherwise}.\\
    \end{cases}
\end{align*}

Therefore, if the centralized control system at some step $t$ adopts an action $A(t)$ for $S(t)$, the next state can be calculated as
\begin{equation*}
S(t + 1) = 
\begin{bmatrix}
x_0(t) + v_{0}(t)\Delta t & f(y_0(t)) & v_{0}(t) + b_{0}(t)\Delta t & g(z_{0}(t), a_{0}(t))\\
x_1(t) + v_{1}(t)\Delta t & f(y_1(t)) & v_{1}(t) + b_{1}(t)\Delta t & g(z_{1}(t), a_{1}(t))\\
& \cdots \\
x_{n-1}(t) + v_{n-1}(t)\Delta t & f(y_{n-1}(t)) & v_{n-1}(t) + b_{n-1}(t)\Delta t & g(z_{n-1}(t), a_{n-1}(t))\\
\end{bmatrix}
.
\end{equation*}

\subsection{Reward}
The reward of this MDP framework is primarily determined by the purpose of the application, establishing the DQJL with the shortest amount of time. Initialize reward for each step as $R(t) = 0$, we penalize the number of non-EMVs appearing on the upper lane for each step, namely, for each non-EMV, if $x_{i} - l_{i} \leq L$ and $y_{i} = 0$, $R(t) \mathrel{{-}{=}} 1$. We determined that the DQJL is established when there are no vehicles remaining on the upper lane.

An important underlying reward definition is that the coordination process must be collision-free, and this can be examined if there are overlapping parts between vehicle's positions. In order to discourage collision, we additionally employ a minimum safety gap $d$ to make sure neighboring vehicles distance themselves. In terms of overlapping between vehicles' positions, $i.e.\ x_{i}(t) + d < x_{j}(t) - l_{j} \text{ and } y_{i}(t) = y_{j}(t)$, a large penalize, $i.e.\ R(t) = -2000$, will be applied for the current step if collision happens.

\subsection{Objective Function}
The objective function for the DQJL problem under MDP framework is
\begin{equation*}
    \max \sum_{t = 0}^{T}R(t),
\end{equation*}
where $T$ stands for the step in which DQJL is established, upper-bounded by $T \leq \frac{L}{v_{b} \times \Delta T}$, where $v_{b}$ is the initial background traffic speed. The constraints are vehicles only have two actions per step, yielding or move forward with original speed.  Solving this objective function not only outputs the optimal coordination strategy, but also reveals the minimum amount of time for a DQJL can be established.

%% file: text/30_Methodology.tex
\section{Methodology}\label{sec:methodology}
In this section, we elaborate how we employ the state-of-the-art deep Q network (DQN) to address the DQJL establishment involving random road environment and stochastic driving behaviors.

After defining the DQJL problem under the MDP framework, it is straightforward to apply model-free Q-learning algorithm to approach the optimal policy. Notice that since the centralized control server is supposed to generate coordination strategy back to RSUs instantaneously, the solution should ensure practicality against all possible road environment such as different number of non-EMVs on the segment, different background traffic speeds and varying vehicle lengths.

\subsection{Q-learning}
In a MDP problem under some policy $\pi$, the combination of some state $s$ and action $a$ will yield a state-action value as
\begin{equation}
    Q_{\pi}(s, a) = E_{\pi}\{\sum_{t = 0}^{K}\{\gamma^{t}R_{t + k + 1}|s_{t} = S, a_{t} = A\}\}.\label{eq:1}
\end{equation}
In \eqref{eq:1}, $E_{\pi}$ represents the expected long term reward under the stochastic policy $\pi$. 
 The $Q_{\pi}(s,a)$ represents the expected long turn reward by the agent in state $s$ choose action $a$ under policy $\pi$. The Q function is represented recursively as:\\
\begin{equation}
\label{eq:2}
    \begin{aligned}
    Q_{\pi}(s, a) & = \sum_{s'}Pr(s'|s, a)( \gamma \sum _{a'}\pi(a'|s')Q_{\pi}(s', a') + R(s, a, s')),
    \end{aligned}
\end{equation}
where $Pr(s'|s, a)$ means the probability of the state collapsed into $s'$ when taking action $a$ in the state $s$, and $R(s, a, s')$ represents the reward for that move.

From \eqref{eq:2}, we can determine the Q function under optimal policy $\pi^{*}$ should satisfy the Bellman's optimality equation:\\
\begin{equation}
    Q^{*}(s, a) = E_{s'}\{R_{t} + \gamma Q^{*}(s', a')\}.\label{eq:optimal}
\end{equation}

When the numbers of states and actions are finite, a simple tabular Q-learning algorithm can be employed to approach convergence through the centralized control system's experience as introduced by \cite{Shota2019Constrained} as
\begin{equation}\label{eq:update}
    Q(s,a) \xleftarrow{} Q(s,a) + \alpha(R_{t+1} + \gamma \max Q(s',a) - Q(s,a)),
\end{equation}
where $0 < \alpha < 1$ represents the learning rate.

Under the traditional Q-learning approach, all non-EMVs may act naively and randomly to take the reward and update the corresponding $Q(s,a)$ by trial-and-error. The centralized control system then plans the next action for the next state based on the collected $Q(s,a)$ and update the new $Q(s',a')$ for the new state and new action. The iterations of the Q-learning will eventually maximize the reward and produce the optimal coordination policy for all non-EMVs at any step. However, the traditional Q-learning algorithm neither deals with the stochastic driving behavior efficiently nor be able to yield coordination strategy in real-time.

\subsection{Deep Q Network}
Since a vehicle's position is continuous in the road environment, the state containing all non-EMVs is also continuous, $i.e.$ there is a infinite number of states. The number of actions is finite and discrete under this MDP framework. Each non-EMV at step $t$ has only 2 actions: continue or yield. The number of actions for the agent, however, depends on the number of non-EMVs in this road segment, $i.e., N_{a} = 2^{n}$, where $n$ is number of non-EMVs. 

To dramatically improve updating frequency, we adopt and modify the state-of-the-art deep Q network (DQN) introduced by \cite{mnih2015humanlevel} to approximate state-action value to output optimal action for each state. Rather than calculating the expected long-term reward by value iterations, DQN approximates state-action value via neural network and has been proven powerful in many real life reinforcement learning application.

\subsubsection{Design of DQN}
To deal with varying numbers of non-EMVs, we incorporate a special technique of padding of the state. That is, we insert additional trivial non-EMVs at the end of the state vector so that the length of state matrix is consistent. Namely, we pad the state vector from $S(t) = [S_{0}(t), S_{1}(t), \dots, S_{n-1}(t)]$ to $S_{p}(t) = [S_{0}(t), s_{1}(t), \dots, S_{n-1}(t), S_{n}(t), \dots, S_{k-1}(t)]$, where $n$ to $k-1$ non-EMVs are trivial vehicles containing meaningless information. 

When $n$ is larger than 10, it also raises concerns that the discrete action space is too large for computation and will harm training performance by making loss functions less reliable when comparing Q-value between states. To cope with large number of possible actions, we down-scale the action space by only allowing no more than one non-EMV to yield during any step. We use a scalar to represent the action adopted by the agent, which is $A(t) = m$ where $m \in \{-1, 0, 1, \dots, n-1\}$ and stands for the vehicle who needs to yield at this step. When $A(t) = -1$, there is no vehicle instructed to yield. When $\Delta t$ is small, this technique becomes feasible as vehicles yielding process can be separated.

At the same time, although vehicle's most comfortable $b_{i}^{*}$ and vehicle length $l_{i}$ are constant during state transition, it is necessary to incorporate them in the state vector so that all features are included in training. 

The trivial non-EMVs, of course, will not interact with other vehicles and each other. They are also ignored when calculating reward for the state. After padding, the state vector has fixed fixed number of non-EMVs, and can be expressed as
\begin{equation*}
S_{p}(t) = 
\begin{bmatrix}
x_0(t) & y_0(t) & v_{0}(t) & z_{0}(t) & b^{*}_{0} & l_{0}\\
x_1(t) & y_1(t) & v_{1}(t) & z_{1}(t) & b^{*}_{1} & l_{1}\\
& \cdots \\
x_{n-1}(t) & y_{n-1}(t) & v_{n-1}(t) & z_{n-1}(t) & b^{*}_{n-1} & l_{n-1}\\
x_{n}(t) & y_{n}(t) & v_{n}(t) & z_{n}(t) & b^{*}_{n} & l_{n}\\
& \cdots \\
x_{k-1}(t) & y_{k-1}(t) & v_{k-1}(t) & z_{k-1}(t) & b^{*}_{k-1} & l_{k-1}\\
\end{bmatrix}
.
\end{equation*}

\begin{figure}[!ht]
  \centering
  \includegraphics[width=\textwidth]{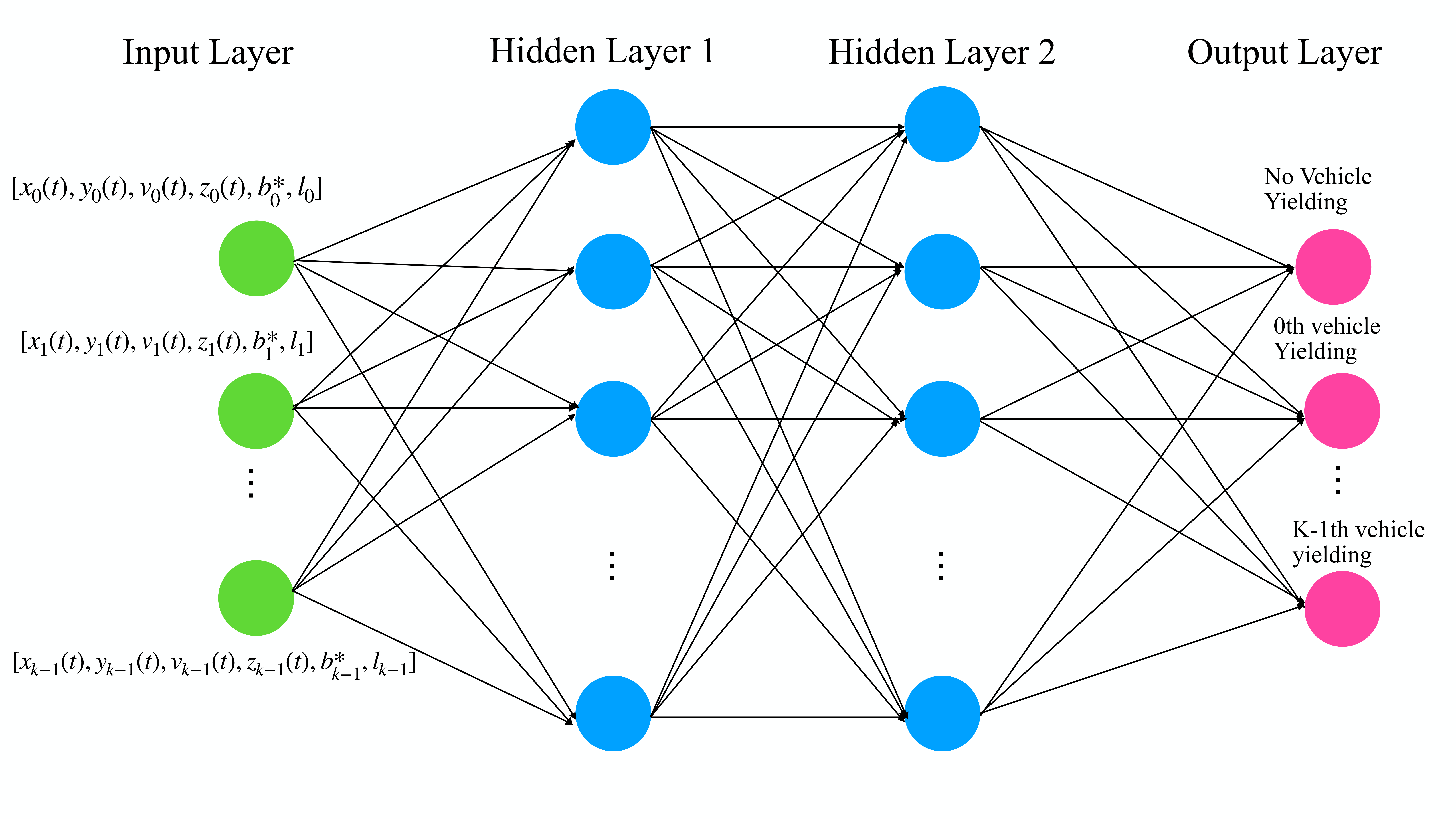}
  \caption{The double hidden layer DQN structure.}
  \label{fig:dqn_structure}
\end{figure}

As the state transition are linear transformations, we employ a multi-layer perceptron (MLP) as the neural network for evaluation and target network. The structure of the neural network is presented as Figure \ref{fig:dqn_structure}. The dimensions of each layer are input layer: $B\text{(batch size)} \times K\text{(number of vehicles)} \times 6\text{(features per vehicle)}$, first hidden layer: $6K \times 128$, second hidden layer: $128 \times 256$ and output layer: $256 \times (K+1) \text{(action size)}$. For each hidden layer, we use ReLU as the activation function.

\subsubsection{Epsilon-greedy Algorithm}
To balance exploration and exploitation, we incorporate $\epsilon$-greedy algorithm to make the agent more and more "confident" to select the the optimal action through training. Mathematically speaking,\\
\begin{equation*}
    A(t) =\begin{cases}
        \argmax Q_{t}(s, a), &\text{with probability of } 1 - \epsilon,\\
        \text{random action}, &\text{with probability of } \epsilon.\\
        \end{cases}
\end{equation*}
As the training proceeds, $\epsilon$ will decay linearly so that the agent will choose action less and less randomly.

\subsubsection{Experience Replay}
For states where the agent has never been, we need an evaluation function to approximate the rewards for those states. However, updating weights of the neural network for a specific pair of state and action will impose change to the $Q(s,a)$ for other pairs of state and action, which may result in significant increase in the training time or even failure to converge. Experience replay is introduced by \cite{Schaul2015PrioritizedER} to store some of experience as a tuple of $(S(t), A(t), R(t), S(t+1))$ into an experience history queue $D$. An off-policy Q-learning algorithm will benefit by randomly select experience tuples with size of the mini-batch from $D$ so that each memory tuple has equal chance to be selected into the training.

\subsubsection{Training DQN with Fixed Q target}
Another important characteristic powering  DQN is the fixed Q target. After every certain training episodes, we replace the weights in the target network by the ones in the evaluation network. Otherwise, we fix the weights in the target network to increase the efficiency of training. 
Originally, DQN updates network parameters by minimizing loss as
\begin{equation*}
    L(\theta_{i}) = E_{s'}{(R + \gamma \max_{a'}Q(s',a';\theta_{i}) - Q(s,a;\theta_{i}))^{2}},\label{eq:loss}
\end{equation*}
where $\theta_{i}$ refers to a specific weight in the neural network and $E$ represents expected long-term reward. Taking the partial derivative with respect to $\theta_{i}$ and we can update network parameters by
\begin{equation*}
\nabla_{\theta_{i}} L(w_{i}) = E_{s'}[R + \gamma max_{a'}Q(s',a';\theta_{i}) - Q(s,a;\theta_{i})\nabla_{\theta_{i}} Q(s,a;\theta_{i})].
\end{equation*}
With fixed Q-target, we calculate the loss in a new way as
\begin{equation*}
    L(\theta_{i}) = E_{s'}[(r + \gamma \max_{a'}Q(s',a';\theta_{i}^{-}) - Q(s,a;\theta_{i}))^{2}],\label{eq:loss_new}
\end{equation*}
where $\theta_{i}^{-}$ is the fixed weight parameter and only gets updated every certain steps. From there, the network performs a stochastic gradient descent to update $\theta_{i}$ and, accordingly, all weights of this neural network. For the optimization process, we employ the Adam optimizer, and the hyper-parameters  selected for training are presented in Table \ref{tab:dqn_parameters}.
\begin{table}[!ht]
	\caption{Hyper-parameters for DQN Training}\label{tab:dqn_parameters}
	\begin{center}
		\begin{tabular}{l l l l}
			Hyper-parameters & Value \\\hline
			Replay Memory Size (D)   & 100,000 \\
			Minibatch Size (B)   & 64 \\
			Starting $\epsilon$   & 1 \\
			Ending $\epsilon$  & 0.001\\
			$\epsilon$ decay steps & 10,000 \\
			Discount factor ($\gamma$) & 0.99 \\
			Learning rate ($\alpha$)   & 0.0005 \\
            Target network update steps & 100\\
            \hline
		\end{tabular}
	\end{center}
\end{table}

\subsection{Generating Yielding Time Indicator}
After training, we apply offline training to output the optimal action sequence for DQJL coordination. For a given initial road environment $S(0)$ with padding, we forward feed the state to get its corresponding optimal action $A^{*}(0)$ and, then, compute the next state $S(1)$. We store $A^{*}(0)$ in to the sequence and repeat the cycle until DQJL has been established. Eventually, we will get $A^{*}$ storing actions for all steps until the end. After dumping actions for no vehicles yielding and for trivial vehicles, we can get a yielding time indicator $T^{*} = [T_{0}, T_{1}, T_{2}, \dots, T_{n-1}]$, where $T_{i}$ is the step index to yield, or $T_{i} = -1$ if the vehicle is not involved in the coordination process at all.

\subsection{Modified DQN Algorithm Summary}
Summarizing above, we propose the modified DQN algorithm to solve the DQJL problem for any given initial road environment. See Algorithm \ref{algo:training} for DQN training and yielding time indicator generation process.\\
\begin{algorithm}[!ht]
\caption{Modified DQN Training and Yielding Time Indicator Generation for DQJL problem}
\begin{algorithmic}[1]\label{algo:training}
\STATE Initialize experience history queue $D$ with mini-batch size
\STATE Initialize evaluation network $Q(s, a)$ with set of weights $\theta$
\STATE Initialize target network $Q(s, a)$ with set of weights $\theta^{-}$
\FOR{coordination training episode}
    \STATE Initialize a random state $S(0) = [S_{0}(0), S_{1}(0), \dots, S_{n-1}(0)]$
    \STATE Include vehicle attributes and pad trivial vehicles into a state $S_{p}(0)$ with fixed length
    \FOR{coordination training step}
        \STATE Select an action $A(t)$ to perform in $S(t)$
        \STATE Update reward $R(t)$ and the next state $S(t+1)$
        \STATE Store the tuple into D as ($S(t), A(t), R(t), S(t+1)$)
        \STATE Collect experience samples ($S(j), A(j), R(j), S(j+1)$) with size of mini-batch
        \STATE Transform ($S(j), A(j), R(j), S(j+1)$) into a training pair ($x_{k}, y_{k}$) by have $x_{k} = S(j)$ and $y_{k} = R(j) + \gamma \max_{A'}Q(S(j+1), A';\theta)$
        \STATE Update $\theta$ for the training pair of ($x_{k}, y_{k}$)
        \STATE Replace $\theta$ by $\theta^{-}$ every few steps 
    \ENDFOR
\ENDFOR

\STATE Extract the target network with trained parameters $\theta$
\STATE Include vehicle attributes and pad trivial vehicles into a state $S_{p}(0)$
\STATE Initialize an optimal action sequence $A^{*}$
\WHILE{vehicle left on upper lane within the road segment}
    \STATE Apply $S(t)$ into the forward-feed network to output $A^{*}(t)$
    \STATE Update $S(t+1)$ based on $S(t)$ and $A^{*}(t)$
    \STATE Store $A^{*}(t)$ into $A^{*}$
\ENDWHILE
\STATE Delete corresponding actions in $A^{*}$ for trivial vehicles
\STATE Generate $T^{*}$ based on $A^{*}$
\end{algorithmic}
\end{algorithm}

\subsection{Double DQN, Dueling DQN and Double Dueling DQN}
To improve and evaluate the training performance of the proposed DQN algorithm, we implement advanced versions of DQN: double DQN (DDQN), dueling DQN, and dueling double DQN (3DQN).

Double DQN consists of two identical neural network models as introduced in \cite{HasseltGS15}. It learns through experience replay through the first model just like DQN does, but it computes Q-value through the secondary model, which is a copy of main model's last episode. Such design effectively handles overestimation issue in DQN as the agent may fail to learn if Q-values are different in experience replay and in actual training. By employing $Q(s,a)$ and $Q'(s,a)$, DDQN separates the process of choosing optimal action and calculating Q-value for that action. So instead of updating Q-value in \eqref{eq:update}, DDQN chooses optimal action by $Q(s,a)$ and estimates expected Q-value by $Q'(s,a)$, which is described as:
\begin{equation*}
    Q(s,a) \xleftarrow{} Q(s,a) + \alpha(R_{t+1} + \gamma Q'(s',a) - Q(s,a)).\label{eq:new_update}
\end{equation*}
The rest remains the same with DQN.

Dueling DQN, introduced by \cite{DBLP:WangFL15}, incorporates the concept of advantage as $A(s, a) = Q(s, a) - V(s)$. The neural network structure also divides into two stream, one of which handles estimating state-value and the other handles estimating all state-dependent advantages value. Aggregating both stream in the practical way, we calculate the Q-value via
\begin{equation*}
    Q(s,a;\theta, \theta_{A}, \theta_{V}) = V(s; \theta, \theta_{V}) + A(s, a; \theta, \theta_{A}) - \frac{1}{\emph{A}}\sum A(s, a'; \theta, \theta_{A}),
\end{equation*}
where $\theta$ stands for common network parameters, $\theta_{A}$ stands for advantage stream parameters and $\theta_{V}$ stands for value stream parameters. Combining double Q-learning and advantage mechanism, we are able to implement 3DQN.
\subsection{Deep Reinforcement Learning Training Result}
As described in the algorithm summary, the training is conducted on simulated scenarios where initial road environments, $i.e.$ initial state vector $S_{0}$, are generated for each episode. When generating vehicle's unique attributes, we use $l_{i} \sim U(4m, 5.5m)$ to capture the length majority private vehicles in road as suggested in \cite{Sellen2019}, and use $b_{i}^{*} \sim N(3.5m/s^{2}, 1m/s^{2})$ to indicate most comfortable deceleration of non-EMVs as suggested in \cite{BOKARE20174733}. Through trial-and-error, the simulated random process in state transition adopts parameters in Table \ref{tab:model_parameters}. With these model parameters, the learning agent precisely simulates the uncertainty in driver' perception reaction time as well as stochastic deceleration behavior. Remember that the perception reaction time is modeled as a geometric distribution with success rate of apply braking in current step $p = \frac{\Delta t}{2.3s}$, and deceleration for non-EMVs performing lane changing is described as $b_{i}(t) \sim N(b_{i}^{*}, \sigma_{0}^{2})$ and $b_{i}(t) \sim N(b_{i}^{*}, \sigma_{1}^{2})$ for non-EMVs braking until stop.
\begin{table}[!ht]
	\caption{Random Process Model Parameters}\label{tab:model_parameters}
	\begin{center}
		\begin{tabular}{l l l l}
			Parameters & Value \\\hline
            Temporal step length ($\Delta t$) & $0.2s$\\
            Road segment length ($L$) & $150m$\\
            State vector length/number of vehicles for training $K$ & 20\\
            Minimum safety gap distance (d) & $0.2m$\\
            Standard deviation for pull-over ($\sigma_{0}$) & $0.8m/s^{2}$\\
            Standard deviation for braking until stop ($\sigma_{1}$) & $0.5m/s^{2}$\\
            \hline
		\end{tabular}
	\end{center}
\end{table}

\begin{figure}[ht]
    \centering
    \begin{subfigure}{0.45\textwidth}
        \centering
        \includegraphics[width=\textwidth]{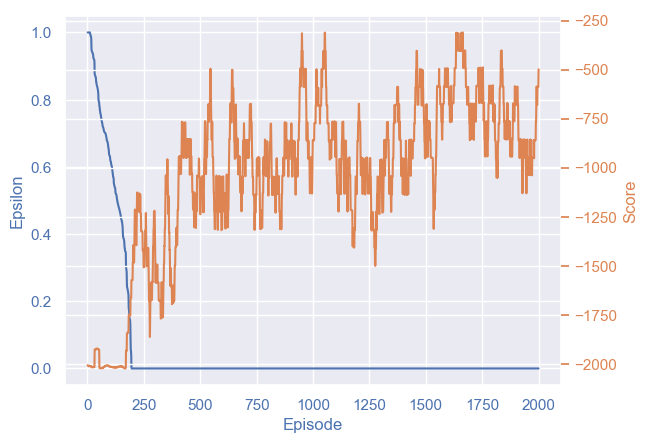}
        \caption[Network2]%
        {{\small DQN training result}}    
        \label{fig:DQN_training_result}
    \end{subfigure}
    \hfill
    \begin{subfigure}{0.45\textwidth}  
        \centering 
        \includegraphics[width=\textwidth]{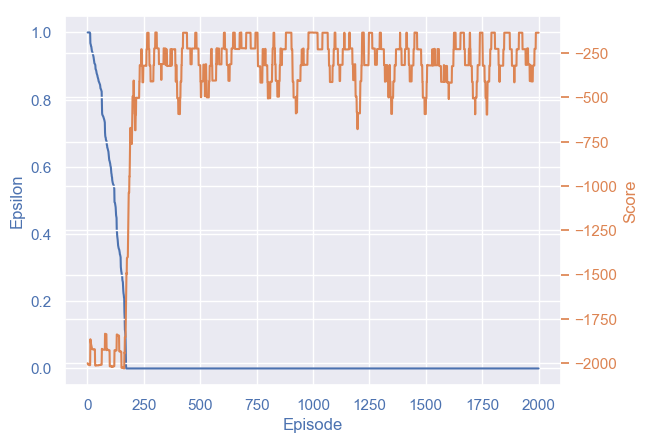}
        \caption[]%
        {{\small DDQN training result}}    
        \label{fig:DDQN_training_result}
    \end{subfigure}
    \vskip\baselineskip
    \begin{subfigure}{0.45\textwidth}   
        \centering 
        \includegraphics[width=\textwidth]{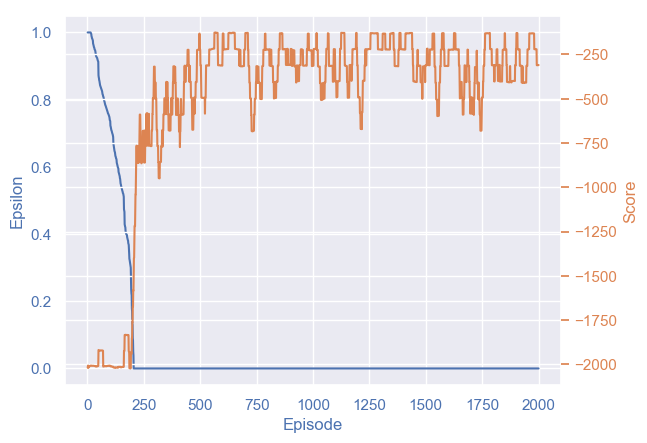}
        \caption[]%
        {{\small Dueling DQN training result}}    
        \label{fig:DuelingDQN_training_result}
    \end{subfigure}
    \hfill
    \begin{subfigure}{0.45\textwidth}   
        \centering 
        \includegraphics[width=\textwidth]{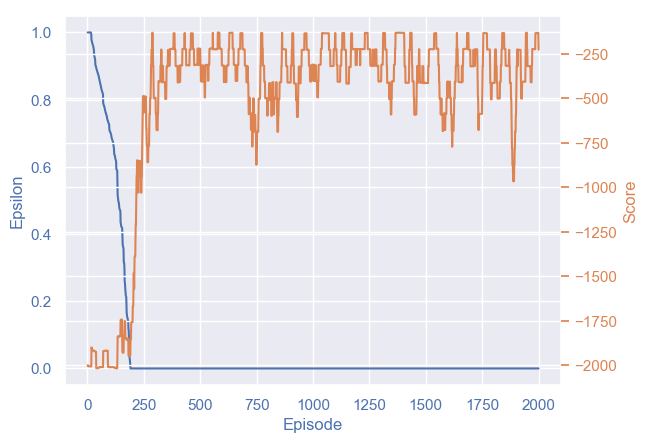}
        \caption[]%
        {{\small 3DQN training result}}    
        \label{fig:DDDQN_training_result}
    \end{subfigure}
    \caption[]
    {\small 2000 Episodes of Training result for DQN, DDQN, Dueling DQN and 3DQN.} 
    \label{fig:training_result}
\end{figure}

The training is completed on an 1.4GHz 4-core Intel Core i5 CPU for all DQN implementation, and it takes about 150 seconds for 2000 episodes of training for each algorithm. The training results are shown in Figure \ref{fig:training_result}. 
\FloatBarrier

As we can see from the learning result in Figure \ref{fig:DQN_training_result}, DQN displays some learning patterns but struggles to converge within 2000 episodes of training. Meanwhile, DDQN and dueling DQN exhibit obvious learning patterns and maintain around an average score of -250 in Figure \ref{fig:DDQN_training_result} and Figure \ref{fig:DuelingDQN_training_result} respectively, and DDQN converges slightly faster than dueling DQN. The fluctuation of the learning curves can be explained by random road environment factors, such as different number of vehicles per episode starts since it's easier to coordinate fewer vehicles than more. The other reason behind the fluctuation is the nature of uncertainty in human driver's perception reaction time and deceleration behavior, so that yielding distance will differ even with same action adopted at same state.

Although converging at the same level, 3DQN presents relatively larger variation in learning as shown in Figure \ref{fig:DDDQN_training_result}. The possible reason behind is implementing both features may "over-correct" the overestimation issue in raw DQN training. By employing double Q-learning and advantage, score difference between actions in some state may be reduced so that agent will choose non-optimal actions.

Based on the learning results, we select the target network in DDQN to generate yielding time indicators with offline training, which can be completed in constant time.

%% file: text/40_Simulation.tex
\section{Simulation-based Validation}\label{sec:simulation}
In this section, we validate our RL-based DQJL coordination strategy on SUMO against the calibrated benchmark system. We also conduct sensitivity analysis against different vehicle densities and background traffic speeds to investigate the most suitable conditions to incorporate our proposed strategy.

\subsection{SUMO Setup and Simulation Implementation}
SUMO, introduced by \cite{SUMO2018}, is a traffic simulation software aiming to precisely simulate real traffic conditions. For this specific application, we define the initial road environment from the randomly generated initial conditions. SUMO simulates the process based on the kinematics of all vehicles, which are determined by SUMO's default car-following model, until EMV has passed this segment. Even though vehicles' velocities and accelerations are now determined by SUMO's car-following model rather than the state transition in the MDP, the stochastic elements such as perception reaction time successfully pick up the dynamic difference on both benchmark system and centralized controlled system. 

To best represent the real life scenarios where non-EMVs are mainly guided by EMV's sirens, we implement the benchmark system where non-EMV's yielding behaviors depend on their distances to the approaching EMV. SUMO's built-in methods $getNeighbors(), slowDown()$, and $changeLane()$ in the $TraCI.\_vehicle$ package manage to let non-EMVs yield when EMV approaches them. Whenever a non-EMV becomes a neighbor vehicle of the passing EMV, it brakes until stop or pulls over, depending on which lane it is on. 

\begin{figure}[!ht]
  \centering
  \includegraphics[width=\textwidth]{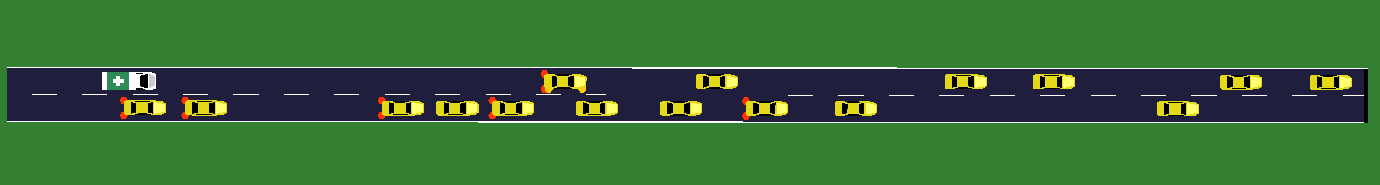}
  \caption{SUMO simulation.}
  \label{fig:sumo}
\end{figure}

To realize the DQJL establishment process with RL-based coordination strategy, we use the yielding time indicator, as the coordination strategy to each randomly generated road environment, to instruct non-EMVs to yield at corresponding timestamps. See Figure \ref{fig:sumo} for a snapshot of SUMO simulation of DQJL process, where vehicles' rear red lights representing they are adjusting their speeds. As the result from RL training, DQJLs are established with absolute safety, showing that the RL-based coordination is feasible not only in the simulation settings but also in real field settings where traffic conditions are more complicated.
 
\FloatBarrier

\subsection{Performance Comparison and  Sensitivity Analysis}
In order to test performance for the benchmark and the RL-based coordination system, we simulate the process of EMV passing through both systems on SUMO. To inspect the impacts on EMV passing time by different factors of the given road environment, we conduct sensitivity analysis on the number of vehicles and background traffic speeds.

\begin{figure}[!ht]
  \centering
  \includegraphics[width=0.7\textwidth]{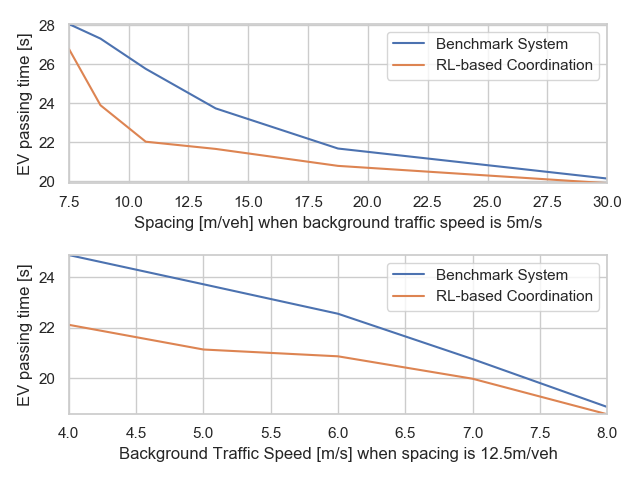}
  \caption{EMV passing time against vehicle density and background traffic speed.}
  \label{fig:sensitivity_analysis}
\end{figure}

Due to the randomness within each scenario, $i.e.$ different starting positions of vehicles when the number of vehicles is fixed, we run 5 experiments for each set and use the average EMV passing time to represent the system performance. When generating initial conditions, vehicle initial positions $x_{i}(0)$ and vehicle unique attributes $l_{i}$ and $b^{*}_{i}$ are totally randomly generated. Road length $L = 150m$ and minimum safety gap $d = 0.2m$ are fixed for all experiments. We also set the maximum allowable speed for the EMV to be $10m/s$, so that EMV can fast run through when no vehicles are blocking its way. The simulation results are shown in Figure \ref{fig:sensitivity_analysis}.

To reflect the impacts of different number of vehicles on EMV's passing time, we calculate the corresponding spacing to better represent congestion levels. It's straightforward to observe that when the average spacing is around 10m/veh, the EMV passing time between two systems differ the most. The result is expected because the RL-based coordination doesn't have advantage over benchmark system when the road environment is too congested or too sparse. 

At the same time, when congestion level is moderate, $i.e.$ an average spacing is 12.5m/veh, we witness the reduction in EMV passing time difference when background traffic speed increases. The pattern can be explained by the time loss experienced by the EMV is minimized when all vehicles are fast, thus EMV can travel with maximum speed for longer period. When all vehicles are running slow, the RL-based coordination strategy manages to save more time to establish the DQJL for EMV. We report the maximum reduction in average EMV passing time with the RL-based coordination to be 14.89\% when background traffic speed is 5m/s and spacing is 12m/veh. Under this setting, the RL-based coordination saves around 2.5 seconds per hundred meters in urban roadway, creating critical time window for EMV to arrive the scene.

%% file: text/50_Conclusion.tex
\section{Conclusion}\label{sec:conclusion}
In this study, we picture the DQJL process under V2I communication technologies and model it under a MDP framework. We propose a modified deep reinforcement learning algorithm to address the DQJL problem, taking into account of stochastic driving behavior and drivers' reaction to approaching EMV. The proposed algorithm is justified effective against any given initial road conditions and can deliver real time coordination strategy under the V2I schema. Our methodology is also justified to outperform benchmark system via validation-based simulation on SUMO, and we inspect the most suitable circumstance to incorporate RL-based coordination strategy.

This study can be extended into few directions. First, we will consider employing multi-agent reinforcement learning for DQJL and other EMV-related connected vehicle applications. Multi-agent RL in the cooperative settings can offer a more nature decomposition of connected vehicle problem and have the potential for more scalable learning. Second, we're extending DQJL process into intersection scenario to formulate intersection-level RL-based coordination strategy. Last but not least, we will incorporate SUMO's car-following models directly into state transition and employ learning-based approach to automate model parameters calibration so that our proposed model will be more extensible and adaptive for real field settings implementation.

\section{Acknowledgements}\label{sec:acknowledgements}
This research was in part supported by NYU Tandon School of Engineering and the C2SMART University Transportation Center.

\section{Author Contributions}\label{sec:author}
The authors confirm contribution to the paper as follows: DQJL concept ideation: J. Y. J. Chow and L. Jin; model formulation and methodology implementation: H. Su and L. Jin; simulation implementation: K. Shi and H. Su; manuscript preparation: H. Su, L. Jin and J. Y. J. Chow. All authors reviewed the results and approved the final
version of the manuscript.

%% file: Su_2020_trb.bbl
\begin{thebibliography}{35}
\providecommand{\natexlab}[1]{#1}

\bibitem[{NY2(2020)}]{NY2019}
\emph{New York End-To-End Response Times}, 2019 (accessed February 28, 2020).

\bibitem[{Eme(2020)}]{Emergency2014}
\emph{Emergency Response Incidents}, 2014 (accessed February 28, 2020).

\bibitem[{Hea(2020)}]{Heart2013}
\emph{Heart Disease and Stroke Statistics}, 2013 (accessed February 28, 2020).

\bibitem[{You et~al.(2020{\natexlab{a}})You, Yang, Chapiro, and
  Duncan}]{you2020unsupervised}
You, C., J.~Yang, J.~Chapiro, and J.~S. Duncan, Unsupervised Wasserstein
  Distance Guided Domain Adaptation for 3D Multi-domain Liver Segmentation. In
  \emph{Interpretable and Annotation-Efficient Learning for Medical Image
  Computing}, Springer International Publishing, 2020{\natexlab{a}}, pp.
  155--163.

\bibitem[{You et~al.(2020{\natexlab{b}})You, Chen, Liu, Yang, and
  Zou}]{you2020data}
You, C., N.~Chen, F.~Liu, D.~Yang, and Y.~Zou, Towards Data Distillation for
  End-to-end Spoken Conversational Question Answering. \emph{arXiv preprint
  arXiv:2010.08923}, 2020{\natexlab{b}}.

\bibitem[{You et~al.(2021{\natexlab{a}})You, Chen, and Zou}]{you2021knowledge}
You, C., N.~Chen, and Y.~Zou, Knowledge distillation for improved accuracy in
  spoken question answering. In \emph{IEEE International Conference on
  Acoustics, Speech and Signal Processing (ICASSP)}, IEEE, 2021{\natexlab{a}},
  pp. 7793--7797.

\bibitem[{Chen et~al.(2021)Chen, Liu, You, Zhou, and Zou}]{chen2021adaptive}
Chen, N., F.~Liu, C.~You, P.~Zhou, and Y.~Zou, Adaptive bi-directional
  attention: Exploring multi-granularity representations for machine reading
  comprehension. In \emph{IEEE International Conference on Acoustics, Speech
  and Signal Processing (ICASSP)}, IEEE, 2021, pp. 7833--7837.

\bibitem[{You et~al.(2020{\natexlab{c}})You, Chen, and
  Zou}]{you2020contextualized}
You, C., N.~Chen, and Y.~Zou, Contextualized Attention-based Knowledge Transfer
  for Spoken Conversational Question Answering. \emph{arXiv preprint
  arXiv:2010.11066}, 2020{\natexlab{c}}.

\bibitem[{You et~al.(2018)You, Yang, Gjesteby, Li, Ju, Zhang, Zhao, Zhang,
  Cong, Wang et~al.}]{you2018structurally}
You, C., Q.~Yang, L.~Gjesteby, G.~Li, S.~Ju, Z.~Zhang, Z.~Zhao, Y.~Zhang,
  W.~Cong, G.~Wang, et~al., Structurally-sensitive multi-scale deep neural
  network for low-dose CT denoising. \emph{IEEE Access}, Vol.~6, 2018, pp.
  41839--41855.

\bibitem[{You et~al.(2019{\natexlab{a}})You, Yang, Zhang, and
  Wang}]{you2019low}
You, C., L.~Yang, Y.~Zhang, and G.~Wang, Low-{D}ose {CT} via {D}eep {CNN} with
  {S}kip {C}onnection and {N}etwork in {N}etwork. In \emph{Developments in
  X-Ray Tomography XII}, International Society for Optics and Photonics,
  2019{\natexlab{a}}, Vol. 11113, p. 111131W.

\bibitem[{You et~al.(2019{\natexlab{b}})You, Li, Zhang, Zhang, Shan, Li, Ju,
  Zhao, Zhang, Cong et~al.}]{you2019ct}
You, C., G.~Li, Y.~Zhang, X.~Zhang, H.~Shan, M.~Li, S.~Ju, Z.~Zhao, Z.~Zhang,
  W.~Cong, et~al., {CT} super-resolution {GAN} constrained by the identical,
  residual, and cycle learning ensemble (GAN-CIRCLE). \emph{IEEE Transactions
  on Medical Imaging}, Vol.~39, No.~1, 2019{\natexlab{b}}, pp. 188--203.

\bibitem[{You et~al.(2021{\natexlab{b}})You, Zhao, Staib, and
  Duncan}]{you2021momentum}
You, C., R.~Zhao, L.~Staib, and J.~S. Duncan, Momentum Contrastive Voxel-wise
  Representation Learning for Semi-supervised Volumetric Medical Image
  Segmentation. \emph{arXiv preprint arXiv:2105.07059}, 2021{\natexlab{b}}.

\bibitem[{Zhou and Gan(2005)}]{Zhou2005Performance}
Zhou, G. and A.~Gan, Performance of Transit Signal Priority with Queue Jumper
  Lanes. \emph{Transportation Research Record}, Vol. 1925, No.~1, 2005, pp.
  265--271.

\bibitem[{Cesme et~al.(2015)Cesme, Altun, and Lane}]{Cesme2015Queue}
Cesme, B., S.~Z. Altun, and B.~Lane, Queue Jump Lane, Transit Signal Priority,
  and Stop Location Evaluation of Transit Preferential Treatments Using
  Microsimulation. \emph{Transportation Research Record}, Vol. 2533, No.~1,
  2015, pp. 39--49.

\bibitem[{Farid et~al.(2015)Farid, Christofa, and Collura}]{Farid2015Dedicated}
Farid, Y.~Z., E.~Christofa, and J.~Collura, Dedicated bus and queue jumper
  lanes at signalized intersections with nearside bus stops.
  \emph{Transportation Research Record: Journal of the Transportation Research
  Board}, Vol. 2484, 2015, pp. 182--192.

\bibitem[{Buchenscheit et~al.(2009)Buchenscheit, Schaub, Kargl, and
  Weber}]{Buchenscheit2009AVE}
Buchenscheit, A., F.~Schaub, F.~Kargl, and M.~Weber, A VANET-based emergency
  vehicle warning system. \emph{2009 IEEE Vehicular Networking Conference
  (VNC)}, 2009, pp. 1--8.

\bibitem[{Yasmin et~al.(2012)Yasmin, Anowar, and Tay}]{Yasmin2012Effects}
Yasmin, S., S.~Anowar, and R.~Tay, Effects of Drivers’ Actions on Severity of
  Emergency Vehicle Collisions. \emph{Transportation Research Record}, Vol.
  2318, No.~1, 2012, pp. 90--97.

\bibitem[{Savolainen et~al.(2010)Savolainen, Datta, Ghosh, and
  Gates}]{Savolainen2010Effects}
Savolainen, P.~T., T.~K. Datta, I.~Ghosh, and T.~J. Gates, Effects of
  Dynamically Activated Emergency Vehicle Warning Sign on Driver Behavior at
  Urban Intersections. \emph{Transportation Research Record}, Vol. 2149, No.~1,
  2010, pp. 77--83.

\bibitem[{Krajzewicz et~al.(2002)Krajzewicz, Hertkorn, R\"ossel, and
  Wagner}]{Krajzewicz2002b}
Krajzewicz, D., G.~Hertkorn, C.~R\"ossel, and P.~Wagner, An Example of
  Microscopic Car Models Validation using the open source Traffic Simulation
  SUMO. In \emph{14th European Simulation Symposium}, 2002, Vol. Jahrgang 2002
  of \emph{SCS European Publishing House}, pp. 318--322,
  lIDO-Berichtsjahr=2004,.

\bibitem[{Zuo et~al.(2019)Zuo, Ozbay, Kurkcu, Gao, Yang, and
  Xie}]{Fan2019Microscopic}
Zuo, F., K.~Ozbay, A.~Kurkcu, J.~Gao, H.~Yang, and K.~Xie, Microscopic
  Simulation based study of Pedestrian Safety Applications at Signalized Urban
  Crossings in a Connected-Automated Vehicle Environment and Reinforcement
  Learning based Optimization of Vehicle Decisions. In \emph{Road Safety and
  Simulation}, 2019.

\bibitem[{Xiong et~al.(2016)Xiong, Wang, Zhang, and Li}]{Xiong2016CombiningDR}
Xiong, X., J.~Wang, F.~Zhang, and K.~Li, Combining Deep Reinforcement Learning
  and Safety Based Control for Autonomous Driving. \emph{ArXiv}, Vol.
  abs/1612.00147, 2016.

\bibitem[{{Schouwenaars} et~al.(2001){Schouwenaars}, {De Moor}, {Feron}, and
  {How}}]{Schouwenaars2001Mixed}
{Schouwenaars}, T., B.~{De Moor}, E.~{Feron}, and J.~{How}, Mixed integer
  programming for multi-vehicle path planning. In \emph{2001 European Control
  Conference (ECC)}, 2001, pp. 2603--2608.

\bibitem[{{Hannoun} et~al.(2019){Hannoun}, {Murray-Tuite}, {Heaslip}, and
  {Chantem}}]{Hannoun2019Facilitating}
{Hannoun}, G.~J., P.~{Murray-Tuite}, K.~{Heaslip}, and T.~{Chantem},
  Facilitating Emergency Response Vehicles’ Movement Through a Road Segment
  in a Connected Vehicle Environment. \emph{IEEE Transactions on Intelligent
  Transportation Systems}, Vol.~20, No.~9, 2019, pp. 3546--3557.

\bibitem[{Santa et~al.(2008)Santa, Gómez-Skarmeta, and
  Sánchez-Artigas}]{SANTA20082850}
Santa, J., A.~F. Gómez-Skarmeta, and M.~Sánchez-Artigas, Architecture and
  evaluation of a unified V2V and V2I communication system based on cellular
  networks. \emph{Computer Communications}, Vol.~31, No.~12, 2008, pp. 2850 --
  2861, mobility Protocols for ITS/VANET.

\bibitem[{Milanés et~al.(2010)Milanés, Godoy, Pérez, Vinagre, González,
  Onieva, and Alonso}]{MILANES201085}
Milanés, V., J.~Godoy, J.~Pérez, B.~Vinagre, C.~González, E.~Onieva, and
  J.~Alonso, V2I-Based Architecture for Information Exchange among Vehicles.
  \emph{IFAC Proceedings Volumes}, Vol.~43, No.~16, 2010, pp. 85 -- 90, 7th
  IFAC Symposium on Intelligent Autonomous Vehicles.

\bibitem[{Muhammad and Safdar(2018)}]{MUHAMMAD201850}
Muhammad, M. and G.~A. Safdar, Survey on existing authentication issues for
  cellular-assisted V2X communication. \emph{Vehicular Communications},
  Vol.~12, 2018, pp. 50 -- 65.

\bibitem[{McGehee et~al.(2000)McGehee, Mazzae, and Baldwin}]{McGehee2000}
McGehee, D.~V., E.~N. Mazzae, and G.~S. Baldwin, Driver Reaction Time in Crash
  Avoidance Research: Validation of a Driving Simulator Study on a Test Track.
  \emph{Proceedings of the Human Factors and Ergonomics Society Annual
  Meeting}, Vol.~44, No.~20, 2000, pp. 3--320--3--323.

\bibitem[{Ohnishi et~al.(2019)Ohnishi, Uchibe, Yamaguchi, Nakanishi, Yasui, and
  Ishii}]{Shota2019Constrained}
Ohnishi, S., E.~Uchibe, Y.~Yamaguchi, K.~Nakanishi, Y.~Yasui, and S.~Ishii,
  Constrained Deep Q-Learning Gradually Approaching Ordinary Q-Learning.
  \emph{Frontiers in Neurorobotics}, Vol.~13, 2019, p. 103.

\bibitem[{Mnih et~al.(2015)Mnih, Kavukcuoglu, Silver, Rusu, Veness, Bellemare,
  Graves, Riedmiller, Fidjeland, Ostrovski, Petersen, Beattie, Sadik,
  Antonoglou, King, Kumaran, Wierstra, Legg, and Hassabis}]{mnih2015humanlevel}
Mnih, V., K.~Kavukcuoglu, D.~Silver, A.~A. Rusu, J.~Veness, M.~G. Bellemare,
  A.~Graves, M.~Riedmiller, A.~K. Fidjeland, G.~Ostrovski, S.~Petersen,
  C.~Beattie, A.~Sadik, I.~Antonoglou, H.~King, D.~Kumaran, D.~Wierstra,
  S.~Legg, and D.~Hassabis, Human-level control through deep reinforcement
  learning. \emph{Nature}, Vol. 518, No. 7540, 2015, pp. 529--533.

\bibitem[{Schaul et~al.(2015)Schaul, Quan, Antonoglou, and
  Silver}]{Schaul2015PrioritizedER}
Schaul, T., J.~Quan, I.~Antonoglou, and D.~Silver, Prioritized Experience
  Replay. \emph{CoRR}, Vol. abs/1511.05952, 2015.

\bibitem[{van Hasselt et~al.(2015)van Hasselt, Guez, and Silver}]{HasseltGS15}
van Hasselt, H., A.~Guez, and D.~Silver, Deep Reinforcement Learning with
  Double Q-learning. \emph{CoRR}, Vol. abs/1509.06461, 2015.

\bibitem[{Wang et~al.(2015)Wang, de~Freitas, and Lanctot}]{DBLP:WangFL15}
Wang, Z., N.~de~Freitas, and M.~Lanctot, Dueling Network Architectures for Deep
  Reinforcement Learning. \emph{CoRR}, Vol. abs/1511.06581, 2015.

\bibitem[{Sellén(2020)}]{Sellen2019}
Sellén, M., \emph{Average Car Length – List of Car Lengths}, 2019 (accessed
  by July, 29 2020).

\bibitem[{Bokare and Maurya(2017)}]{BOKARE20174733}
Bokare, P. and A.~Maurya, Acceleration-Deceleration Behaviour of Various
  Vehicle Types. \emph{Transportation Research Procedia}, Vol.~25, 2017, pp.
  4733 -- 4749, world Conference on Transport Research - WCTR 2016 Shanghai.
  10-15 July 2016.

\bibitem[{Lopez et~al.(2018)Lopez, Behrisch, Bieker-Walz, Erdmann,
  Fl{\"o}tter{\"o}d, Hilbrich, L{\"u}cken, Rummel, Wagner, and
  Wie{\ss}ner}]{SUMO2018}
Lopez, P.~A., M.~Behrisch, L.~Bieker-Walz, J.~Erdmann, Y.-P. Fl{\"o}tter{\"o}d,
  R.~Hilbrich, L.~L{\"u}cken, J.~Rummel, P.~Wagner, and E.~Wie{\ss}ner,
  Microscopic Traffic Simulation using SUMO. In \emph{The 21st IEEE
  International Conference on Intelligent Transportation Systems}, IEEE, 2018.

\end{thebibliography}
